\documentclass{article}

\usepackage{arxiv}

\usepackage{amsmath,amsfonts,bm}

\def\eqref#1{equation~\ref{#1}}

\def\1{\bm{1}}

\def\vmu{{\bm{\mu}}}
\def\vtheta{{\bm{\theta}}}

\def\vk{{\bm{k}}}

\def\vx{{\bm{x}}}
\def\vy{{\bm{y}}}

\def\mI{{\bm{I}}}

\def\mK{{\bm{K}}}

\def\mX{{\bm{X}}}

\DeclareMathAlphabet{\mathsfit}{\encodingdefault}{\sfdefault}{m}{sl}
\SetMathAlphabet{\mathsfit}{bold}{\encodingdefault}{\sfdefault}{bx}{n}

\def\sR{{\mathbb{R}}}

\DeclareMathOperator*{\argmax}{arg\,max}
\DeclareMathOperator*{\argmin}{arg\,min}

\usepackage{amssymb}
\usepackage{algorithm}
\usepackage[noend]{algpseudocode}
\usepackage{enumerate}   

\usepackage[utf8]{inputenc} 
\usepackage[T1]{fontenc}    
\usepackage{hyperref}       
\usepackage{url}            
\usepackage{booktabs}       
\usepackage{amsfonts}       
\usepackage{nicefrac}       
\usepackage{microtype}      
\usepackage{cleveref}       
\usepackage{lipsum}         
\usepackage{graphicx}
\usepackage{natbib}
\usepackage{doi}

\title{Bayesian Optimisation Using Product-of-Experts Gaussian Process Models with Uncertainty Calibration}

\date{}

\newif\ifuniqueAffiliation
\uniqueAffiliationtrue

\ifuniqueAffiliation 
\author{ \href{https://orcid.org/0000-0002-8452-0999}{\includegraphics[scale=0.06]{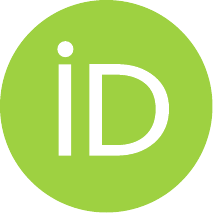}\hspace{1mm}Yean Hoon Ong} \\
	University College London\\
	London, United Kingdom \\
	\texttt{yean-hoon.ong@ucl.ac.uk} \\
}

\renewcommand{\headeright}{}
\renewcommand{\undertitle}{}
\renewcommand{\shorttitle}{Bayesian optimisation using product-of-experts Gaussian process models}

\hypersetup{
pdftitle={Bayesian Optimisation Using Product-of-Experts Gaussian Process Models with Uncertainty Calibration},
pdfsubject={cs.LG},
pdfauthor={Yean Hoon Ong},
pdfkeywords={Bayesian optimisation, Product-of-experts Gaussian processes, Uncertainty calibration},
}

\begin{document}
\maketitle

\begin{abstract} 
Bayesian optimisation (BO) typically relies on a single global Gaussian process (GP) model as its surrogate model. However, GP regression has cubic computational complexity in the number of training data points, limiting its applicability to large-scale optimisation problems. 
The product-of-experts Gaussian process model with uncertainty calibration (GP-pro-c) mitigates this limitation by combining multiple local GP experts, enabling improved uncertainty quantification, reduced computational cost, and preservation of global correlations. 
Despite these desirable properties, the use of GP-pro-c in BO has not been thoroughly studied. 
This paper introduces BO-pro-c, a Bayesian optimisation algorithm that uses GP-pro-c as its surrogate model, and evaluate its performance across a diverse range of BO settings. 
Experimental results suggest that BO-pro-c maintains competitive optimisation performance while achieving a 0.9\% reduction in simple regret and a 39.4\% reduction in computational overhead relative to a BO algorithm based on a single global GP model. \end{abstract}


\section{Introduction}
\label{sec:sec_intro}

Bayesian optimisation (BO) \citep{Brochu2010, Shahriari2016p, Frazier2018, garnett_bayesoptbook_2023} is an effective strategy for finding the optimum (maximum or minimum) of a black-box function. Gaussian processes (GPs) are the most commonly used surrogate models in BO because of their ability to provide both predictive means and uncertainty quantification. However, modelling black-box functions effectively with GPs in BO remains challenging.

Traditionally, most BO algorithms employ a single global GP (GP-glo) as the surrogate model. However, GP-glo has a key limitation: training and inference in GP regression require all input data points and involve inversion of an $n \times n$ covariance matrix, where $n$ is the number of training data points \citep{Rasmussen2006, Shahriari2016p}. Consequently, the computational cost scales cubically, i.e., $\mathcal{O}(n^3)$, making GP-glo impractical for large-scale datasets and high-dimensional problems.

To address this limitation, previous studies have proposed the use of collections of independent local GPs (GP-ind) \citep{Eriksson2019} and product-of-experts GP models (GP-pro) \citep{Schilling2016, Tautvaisas2022} as surrogate models. Compared with GP-glo and GP-ind, GP-pro models, which consist of multiple collaborative local GPs, can alleviate the cubic computational burden while capturing both local patterns and global correlations. However, the use of GP-pro as a surrogate model in BO remains limited, and existing approaches often deliver less satisfactory performance than BO algorithms based on GP-glo or GP-ind.

This paper adopts the product-of-experts GP model with uncertainty calibration proposed by \citep{Ong2026}. This model, referred to as GP-pro-c, combines multiple local GP models whose predictions are aggregated to produce a final estimate. The GP-pro-c model offers several advantages: 
(i) it captures local patterns by allowing each local GP to learn region-specific hyperparameters; 
(ii) it captures global spatial correlations; 
(iii) it mitigates cubic computational costs; and 
(iv) it incorporates an information-based mechanism for calibrating overestimated predictive variances. 
Despite these advantages, no comprehensive study has evaluated the effectiveness of GP-pro-c in BO settings across different acquisition functions.

The main contributions of this paper are as follows: 
\begin{itemize} 
   \item Introduction of a BO algorithm that uses GP-pro-c as its surrogate model, denoted as BO-pro-c. 
   \item Comprehensive comparative studies of optimisation performance and computational overhead across a range of BO settings using four widely adopted acquisition functions: expected improvement (EI), Gaussian process lower confidence bound (GP-LCB), max-value entropy search (MES), and Thompson sampling (TS). 
   \item A rigorous evaluation of BO-pro-c against BO algorithms based on GP-pro, GP-glo, and GP-ind models. 
\end{itemize}

The remainder of this paper is organised as follows. 
Section~\ref{sec:sec_s2_related_work} reviews related work. 
Section~\ref{sec:sec_c4_algo} presents the BO-pro-c algorithm. 
Section~\ref{sec:sec_c4_experiments} reports the experimental results, and Section~\ref{sec:sec_s4_conclusion} concludes the paper.

\section{Related Work}
\label{sec:sec_s2_related_work}

The use of generalised product-of-experts Gaussian processes (gPoE) as surrogate models in Bayesian optimisation (BO) has been explored only by \cite{Schilling2016} and \cite{Tautvaisas2022}, and remains relatively underdeveloped.

\cite{Schilling2016} were the first to investigate the use of a gPoE surrogate model in a sequential BO framework for tuning the hyperparameters of machine learning models via a meta-learning approach. 
In their method, each GP expert was assigned an equal number of training data points through random partitioning. Posterior predictions were aggregated using a uniform weighting scheme, whereby each GP expert was assigned the same weight. 
Empirical results obtained using the expected improvement (EI) acquisition function showed that the gPoE-based BO algorithm only occasionally outperformed a BO algorithm based on a single global GP model.

\cite{Tautvaisas2022} extended this line of research by investigating three PoE-based models, namely gPoE, the Bayesian committee machine (BCM), and the robust Bayesian committee machine (rBCM), as surrogate models. 
They also proposed a BO algorithm that combines the gPoE model with a trust-region framework. 
Although entropy-based weighting was introduced to improve posterior aggregation, the approach continued to rely on random data partitioning, which does not explicitly account for spatial structure or local data density. 
Furthermore, the empirical evaluation was restricted to a narrow BO setting, using sequential query selection and the upper confidence bound (UCB) acquisition function with a fixed parameter $\kappa = 1$. 
This limited experimental setting makes it difficult to assess the robustness and general applicability of PoE-based surrogate models in BO.

More importantly, both studies reported that gPoE-based BO algorithms failed to consistently outperform GP-glo or strong baselines such as TuRBO. 
A key issue identified by \cite{Tautvaisas2022} is the overestimation of predictive variances in gPoE models, which can lead to excessive exploration and consequently degrade optimisation performance. 
However, this limitation was not addressed algorithmically. 
Instead, heuristic adjustments, such as fixing $\kappa = 1$, were used to partially mitigate over-exploration, without resolving the underlying modelling issue.

To address this limitation, \cite{Ong2026} proposed the GP-pro-c model, which incorporates an information-theoretic calibration mechanism for correcting overestimated posterior variances in product-of-experts GP models. 
By exploiting the monotonicity and submodularity properties of information gain, the method defines a principled calibration ratio for adjusting the predictive uncertainty of each local expert. This leads to improved uncertainty quantification while preserving predictive accuracy and computational efficiency.

Despite these advances, the potential of calibrated PoE models in BO remains largely unexplored. 
In particular, there has been no systematic investigation of 
(i) whether uncertainty calibration improves optimisation performance, 
(ii) how calibrated PoE models behave under different acquisition functions, and 
(iii) whether such models can provide a practical alternative to GP-glo and existing scalable BO methods.

This paper addresses these gaps by integrating the GP-pro-c model into a BO framework and conducting a comprehensive evaluation across a diverse range of BO settings.

\section{Algorithm BO-pro-c}
\label{sec:sec_c4_algo}

Algorithm~\ref{alg:gp_poe_c} presents the pseudo-code of BO-pro-c, a Bayesian optimisation algorithm that uses a product of $M > 1$ local GP models (GP-pro-c) to model an unknown objective function. An implementation of BO-pro-c is available at \url{https://github.com/yhong123/bo-pro-c}.

\begin{algorithm} 
\caption{BO-pro-c} 
\label{alg:gp_poe_c} 
\begin{algorithmic} 
	\State \textbf{Input:} unknown noisy objective function $f$ 
	\State \textbf{Input:} input domain $\mathcal{X}$ 
	\State \textbf{Input:} initial sample size $N_{\mathrm{init}}$ 
	\State \textbf{Input:} maximum number of data points per local GP model $N_{\mathrm{GP}}$ 
	\State \textbf{Input:} acquisition function $a$ 
	\State \textbf{Input:} number of local GP models $M$ (minimum: $2$) 
	\State \textbf{Input:} evaluation budget $T$
	\State \textbf{Initialise:} $\mathcal{D}_0 = \{(\vx_i,y_i)\}_{i \leq N_{\mathrm{init}}}$
	
	\For{$t = 1,\ldots,T$} 
		\State $M \gets \max\!\left(2,\mathrm{int}(|\mathcal{D}_{t-1}|/N_{\mathrm{GP}})\right)$ 
		\State Partition $\mathcal{D}_{t-1}$ into $M$ subsets $\mathcal{D}^{(m)}_{t-1}$ 
		\State Assign each subset $\mathcal{D}^{(m)}_{t-1}$ to local model $GP^{(m)}_{t-1}$
		\State Train each $GP^{(m)}_{t-1}$ using $\mathcal{D}^{(m)}_{t-1}$ 
		\State Generate candidate set $\mathcal{X}'=\{\vx'_i\}$ 
		\ForAll{$\vx'_i \in \mathcal{X}'$} 
			\State Compute calibration ratios using Eq.~\ref{eq:eq_cc_ratio} 
			\State Compute aggregated posterior mean and variance using Eqs.~\ref{eq:eq_agg_mu}--\ref{eq:eq_agg_sigma} 
		\EndFor
		\State Select the next query point: 
		\begin{align} 
		\vx_t = 
			\begin{cases} 
			\argmax_{\vx'} \; a^{\mathrm{EI}}(\vx' \mid \mathcal{D}_{t-1}) \\ 
			\argmin_{\vx'} \; a^{\mathrm{LCB}}(\vx' \mid \mathcal{D}_{t-1}) \\ 
			\argmax_{\vx'} \; a^{\mathrm{MES}}(\vx' \mid \mathcal{D}_{t-1}) \\ 
			\argmin_{\vx'} \; a^{\mathrm{TS}}(\vx' \mid \mathcal{D}_{t-1}) 
			\end{cases} 
		\end{align}
		\State Evaluate the objective function: $y_t = f(\vx_t) + \epsilon_t$
		\State Update the dataset: $\mathcal{D}_t \gets \mathcal{D}_{t-1} \cup \{(\vx_t,y_t)\}$
	\EndFor 
	\State \textbf{return} $\hat{\vx}_T=\argmin_{\vx\in\mathcal{D}_T} y(\vx)$ 
\end{algorithmic} 
\end{algorithm}

Given an unknown noisy objective function $f : \mathcal{X} \rightarrow \sR$, where $\mathcal{X}$ is a compact subset of $\sR^d$, the objective is to identify the minimiser $\vx_*$ within the input domain $\mathcal{X}$ under a fixed evaluation budget $T$.

\paragraph{Initialisation} \hfill

The BO-pro-c algorithm begins by constructing an initial dataset $\mathcal{D}_{0} = \{(\vx_i, y_i)\}_{i \leqslant N_{init}}$. The initial input locations are generated using Latin hypercube sampling \citep{Jones1998, Bull2011}. Following standard practice in Bayesian optimisation, the initial sample size is set to $N_{init}=2(d+1)$ \citep{balandat2020botorch}.

\paragraph{Data assignment and local GP training} \hfill

Each local GP model is constrained to a maximum of $N_{GP}=200$ data points. 
At iteration $t$, given dataset $\mathcal{D}_{t-1}$ containing $n$ observations, the number of experts is defined as 
\[ 
M = \mathrm{int}(|\mathcal{D}_{t-1}| / N_{GP}), 
\] 
subject to a minimum of $M=2$. 
The dataset is then partitioned into $M$ subsets, $\mathcal{D}^{(m)}_{t-1}$, such that each subset contains approximately the same number of data points. 

Three data assignment strategies are considered: 
(i) random partitioning, 
(ii) $k$-means clustering, and 
(iii) ball-tree partitioning 
(see \cite{Ong2026} for details). 

Each local model $GP_{t-1}^{(m)}$ is trained on $\mathcal{D}_{t-1}^{(m)}$ using a constant mean function and a Matérn-$5/2$ kernel with automatic relevance determination (ARD). The hyperparameters $\vtheta^{(m)}_{t-1}$ are estimated by maximising the log marginal likelihood:
\begin{align}
\log p \left( \vy_{t-1}^{(m)} \mid \mX_{t-1}^{(m)},  \vtheta_{t-1}^{(m)} \right) 
= &- \frac{1}{2} \left( \vy_{t-1}^{(m)} - \tilde{\vmu}_{t-1}^{(m)} \right)^\top 
\left( \mK_{t-1}^{(m)} + \ddot{\sigma}_{t-1,m}^2 \ \mI \right)^{-1} 
\left( \vy_{t-1}^{(m)} -  \tilde{\vmu}_{t-1}^{(m)} \right)  \notag \\
&-  \frac{1}{2} \log \left| \mK_{t-1}^{(m)} + \ddot{\sigma}_{t-1,m}^2 \ \mI \right| 
- \frac{ \left| \mathcal{D}_{t-1}^{(m)} \right| }{2} \log (2\pi)  
\end{align}
where 
$\tilde{\vmu}^{(m)}_{t-1}$ 
is the vector of mean-function values evaluated at 
$\mX^{(m)}_{t-1}$, 
$\ddot{\sigma}_{t-1,m}^{2}$ 
is the observation noise variance, and 
$\mK_{t-1}^{(m)} + \ddot{\sigma}_{t-1,m}^{2}\mI$ 
is the covariance matrix associated with the noisy observations 
$\vy_{t-1}^{(m)}$.

\paragraph{Posterior aggregation with calibration} \hfill

A candidate set $\mathcal{X}'=\{\vx'_i\}$ of size $\min(100dM, 10^4)$ is generated using a scrambled Sobol sequence. 
For each candidate point $\vx'_i$, every local model $GP^{(m)}_{t-1}$ independently computes the posterior predictive mean $\mu_{t-1,m}(\vx'_i)$ and variance $\sigma^2_{t-1,m}(\vx'_i)$:
\begin{align}
\mu_{t-1,m}(\vx'_i) &=  \tilde{\mu}_{t-1}^{(m)}(\vx'_i) +  \vk_{t-1}^{(m)}(\vx'_i)^{\top} \left(  \mK_{t-1}^{(m)} + \ddot{\sigma}_{t-1,m}^2 \ \mI \right)^{-1}  \left( \vy_{t-1}^{(m)} - \tilde{\vmu}_{t-1}^{(m)}  \right) \\
\sigma_{t-1,m}^2(\vx'_i) &= k_{t-1}^{(m)}(\vx'_i, \vx'_i) -  \vk_{t-1}^{(m)}(\vx'_i)^{\top}  \left(  \mK_{t-1}^{(m)} + \ddot{\sigma}_{t-1,m}^2 \ \mI  \right)^{-1} \vk_{t-1}^{(m)}(\vx'_i) 
\end{align}
where $\tilde{\mu}_{t-1}^{(m)}$ denotes the mean function, $k_{t-1}^{(m)}$ denotes the covariance function, and $\vk_{t-1}^{(m)}$ contains the covariance terms between $\vx'_i$ and $\mX_{t-1}^{(m)}$.

To mitigate variance overestimation, the predictive variance of each local model is adjusted using a calibration factor $\alpha_{t-1,m}(\vx'_i)$:
\begin{align}
\hat{\sigma}^2_{t-1,m}(\vx'_i) = \alpha_{t-1,m}(\vx'_i)\,\sigma^2_{t-1,m}(\vx'_i)   \label{eq:eq_cc_ratio}
\end{align}
where
\begin{align}
\alpha_{t-1,m}(\vx'_i) = \exp\!\left(
- \frac{e-1}{e} \ \frac{1}{2} \log\!\left(1 + \frac{\sigma^2_{t-1,m}(\vx'_i)}{\ddot{\sigma}^2_{t-1,m}}\right)
\right)
\end{align}
See \cite{Ong2026} for the derivation of the calibration factor.

The aggregated predictive distribution is given by:
\begin{align}
\mu_{t-1}(\vx'_i) &= \sigma_{t-1}^2(\vx'_i)  \sum_{m=1}^M \left( w_{t-1,m}(\vx'_i) \  \mu_{t-1,m}(\vx'_i) \ \frac{1}{ \hat{\sigma}_{t-1,m}^2(\vx'_i)} \right)   \label{eq:eq_agg_mu}    \\
\sigma_{t-1}^2(\vx'_i) &= \left( \sum_{m=1}^M \left( w_{t-1,m}(\vx'_i) \  \frac{1}{ \hat{\sigma}_{t-1,m}^2(\vx'_i)}  \right)  \right)^{-1}  \label{eq:eq_agg_sigma}
\end{align}
where $w_{t-1,m}(\vx'_i) > 0$ denotes the weight assigned to the $m$-th local model. The weights are computed using either the entropy-based or variance-based weighting scheme described in \cite{Ong2026}.

\paragraph{Acquisition and optimisation} \hfill

The aggregated predictive mean and variance are used to evaluate the acquisition function. Four widely used acquisition functions are considered: expected improvement (EI), Gaussian process lower confidence bound (GP-LCB), max-value entropy search (MES), and Thompson sampling (TS).

The selected query point $\vx_t$ is evaluated to obtain $y_t = f(\vx_t) + \epsilon_t$, and the dataset is updated accordingly. This process is repeated until the evaluation budget $T$ is exhausted.
The final recommendation is 
\[ 
\hat{\vx}_T = \arg\min_{\vx \in \mathcal{D}_T} y(\vx)
\]

\section{Experiments}
\label{sec:sec_c4_experiments}

This section presents the results and discussion of empirical evaluations of BO-pro-c with respect to 
(i) three data assignment methods, 
(ii) three weighting schemes, 
(iii) the use of uncertainty calibration, 
(iv) four acquisition functions, and 
(v) 12 input dimensionalities. 

For clarity, all experiments followed a common evaluation protocol, summarised below.

\textbf{BO algorithms}. 
The experiments compared BO-pro-c against 
(i) BO algorithms based on a single global GP model (BO-glo), 
(ii) BO algorithms based on a collection of independent local GP models (BO-ind), and 
(iii) BO algorithms based on a product-of-experts GP model without uncertainty calibration (BO-pro).

In total, 88 BO algorithm variants were considered: 
\begin{enumerate}[(i)] 
\item four BO-glo variants using the EI, GP-LCB, MES, and TS acquisition functions, respectively; 
\item 12 BO-ind variants corresponding to the Cartesian product of three data assignment methods and four acquisition functions; 
\item 36 BO-pro variants corresponding to the Cartesian product of three data assignment methods, three weighting schemes, and four acquisition functions; and 
\item 36 BO-pro-c variants corresponding to the Cartesian product of three data assignment methods, three weighting schemes, uncertainty calibration, and four acquisition functions. 
\end{enumerate}
All BO-ind, BO-pro, and BO-pro-c variants were constrained to a maximum of 200 data points per local GP model.

Table~\ref{tab:tab_c4_gp_models} summarises the BO algorithm variants considered in the experiments. 
The algorithm names (e.g., \texttt{pro-rd-ent-c}) encoded the corresponding model configurations. 
The labels \texttt{glo}, \texttt{ind}, and \texttt{pro} denoted the single global GP model, independent local GP models, and product-of-experts GP models, respectively. 
The labels \texttt{rd}, \texttt{kx}, and \texttt{bt} denoted random, $k$-means, and ball-tree data assignment methods, respectively, whereas \texttt{ent}, \texttt{var}, and \texttt{uni} denoted entropy, variance, and uniform weighting schemes. 
The suffix \texttt{c} indicated the use of uncertainty calibration.

\begin{table}
\centering
\caption{Summary of BO algorithm variants used in the experiments, including BO-pro-c, BO-pro, BO-ind, and BO-glo.}
\begin{footnotesize}
\begin{tabular}[t]{lllllll}
\toprule
Label & Type of GPs  &  Data assignment     &  Weighting    &  Uncertainty   &  \\
	        &                        &     method         &   scheme      &  calibration & \\
\midrule
glo                  &  single global GP         &       &         \\
ind-rd         &   independent local GPs            &  random     &       \\
ind-kx         &   independent local GPs          &  $k$-means      &       \\
ind-bt         &   independent local GPs            &  ball-tree     &       \\
\vspace{1\baselineskip}\\ 
pro-bt-ent         &  product-of-experts GP            &  ball-tree     &   entropy          & no       \\
pro-bt-var         &  product-of-experts GP            &  ball-tree      &   variance           & no      \\
pro-bt-uni         &  product-of-experts GP            &  ball-tree    &   uniform          & no       \\
\vspace{1\baselineskip}\\ 
pro-kx-ent         &  product-of-experts GP            &  $k$-means     &   entropy          & no       \\
pro-kx-var         &  product-of-experts GP            &  $k$-means      &   variance           & no      \\
pro-kx-uni         &  product-of-experts GP            &  $k$-means     &   uniform          & no       \\
\vspace{1\baselineskip}\\ 
pro-rd-ent         &  product-of-experts GP            &  random     &   entropy         & no        \\
pro-rd-var         &  product-of-experts GP            &  random     &   variance         & no        \\
pro-rd-uni         &  product-of-experts GP            &  random     &   uniform         & no        \\
\vspace{1\baselineskip}\\ 
pro-bt-ent-c         &  product-of-experts GP            &  ball-tree     &   entropy         & yes        \\
pro-bt-var-c         &  product-of-experts GP            &  ball-tree     &   variance         & yes        \\
pro-bt-uni-c         &  product-of-experts GP            &  ball-tree     &   uniform         & yes        \\
\vspace{1\baselineskip}\\ 
pro-kx-ent-c         &  product-of-experts GP            &  $k$-means     &   entropy          & yes       \\
pro-kx-var-c         &  product-of-experts GP            &  $k$-means      &   variance           & yes      \\
pro-kx-uni-c         &  product-of-experts GP            &  $k$-means     &   uniform          & yes       \\
\vspace{1\baselineskip}\\ 
pro-rd-ent-c         &  product-of-experts GP            &  random     &   entropy         & yes        \\
pro-rd-var-c         &  product-of-experts GP            &  random     &   variance         & yes        \\
pro-rd-uni-c         &  product-of-experts GP            &  random     &   uniform         & yes        \\
\bottomrule
\label{tab:tab_c4_gp_models}
\end{tabular}
\end{footnotesize}
\end{table}

\textbf{GP modelling}. 
All experiments used GPyTorch \citep{Gardner2018} for GP regression. Before model fitting, the input domain was rescaled to $[0,1]^d$ and the function values were standardised. Each GP employed a Matérn-$5/2$ covariance function with automatic relevance determination (ARD) and a constant mean function. The GP hyperparameters were estimated by maximising the log marginal likelihood prior to selecting new query points.

\textbf{Synthetic benchmark functions}. 
Experiments were conducted on 47 benchmark functions with known global minima: the two-dimensional Eggholder, Goldstein-Price, and Shubert functions, together with the Ackley, Levy, Rastrigin, and Rosenbrock functions in dimensions $d=5,\ldots,15$. The batch size was set to $B=d$. 
Table~\ref{tab:tab_c4_bo_benchmark_fns} summarises the benchmark functions. 
All objective function evaluations were corrupted with additive Gaussian noise drawn from $\mathcal{N}(0,0.25)$.

\begin{table}
\centering
\caption{Summary of the 47 synthetic benchmark functions used in the BO experiments.}
\begin{footnotesize}
\begin{tabular}[t]{llllll}
\toprule
Function & Number of &  Input    &  Minimiser   &  Global minimum  &  \\
	      &  variables  & domain  & $\vx_*$       & $f(\vx_*)$ & \\
\midrule
Eggholder                  &  2          &  $[-512, 512]^2$       &  (512, 404.2319)     & -959.6407       \\
Goldstein-Price          &  2          &  $[-2, 2]^2$       &  (0, -1)     & 3       \\
Shubert                      &  2          &  $[-10, 10]^2$       &  multiple     & -186.7309       \\

Ackley          &  5          &  $[-10, 10]^5$       &  (0, 0, 0, 0, 0)     & 0      \\
Levy          &  5          &  $[-10, 10]^5$       &  (1, 1, 1, 1, 1)     & 0      \\
Rastrigin          &  5          &  $[-5.12, 5.12]^5$       &  (0, 0, 0, 0, 0)     & 0      \\
Rosenbrock          &  5          &  $[-5, 10]^5$       &  (1, 1, 1, 1, 1)     & 0      \\

Ackley          &  6          &  $[-10, 10]^6$       &   (0, $\dots$, 0)     & 0      \\
Levy          &  6          &  $[-10, 10]^6$       &  (1, $\dots$, 1)     & 0      \\
Rastrigin          &  6          &  $[-5.12, 5.12]^6$       &   (0, $\dots$, 0)     & 0      \\
Rosenbrock          &  6          &  $[-5, 10]^6$       &  (1, $\dots$, 1)    & 0      \\

Ackley          &  7          &  $[-10, 10]^7$       &   (0, $\dots$, 0)     & 0      \\
Levy          &  7          &  $[-10, 10]^7$       &  (1, $\dots$, 1)     & 0      \\
Rastrigin          &  7          &  $[-5.12, 5.12]^7$       &   (0, $\dots$, 0)     & 0      \\
Rosenbrock          & 7          &  $[-5, 10]^7$       &  (1, $\dots$, 1)    & 0      \\

Ackley          &  8          &  $[-10, 10]^8$       &   (0, $\dots$, 0)     & 0      \\
Levy          &  8          &  $[-10, 10]^8$       &  (1, $\dots$, 1)     & 0      \\
Rastrigin          & 8          &  $[-5.12, 5.12]^8$       &   (0, $\dots$, 0)     & 0      \\
Rosenbrock          &  8          &  $[-5, 10]^8$       &  (1, $\dots$, 1)    & 0      \\

Ackley          &  9          &  $[-10, 10]^9$       &   (0, $\dots$, 0)     & 0      \\
Levy          &  9          &  $[-10, 10]^9$       &  (1, $\dots$, 1)     & 0      \\
Rastrigin          &  9          &  $[-5.12, 5.12]^9$       &   (0, $\dots$, 0)     & 0      \\
Rosenbrock          &  9          &  $[-5, 10]^9$       &  (1, $\dots$, 1)    & 0      \\

Ackley          &  10          &  $[-10, 10]^{10}$       &   (0, $\dots$, 0)     & 0      \\
Levy          &  10          &  $[-10, 10]^{10}$       &  (1, $\dots$, 1)     & 0      \\
Rastrigin          &  10          &  $[-5.12, 5.12]^{10}$       &  (0, $\dots$, 0)     & 0      \\
Rosenbrock          &  10          &  $[-5, 10]^{10}$       &  (1, $\dots$, 1)      & 0      \\

Ackley          &  11          &  $[-10, 10]^{11}$       &   (0, $\dots$, 0)     & 0      \\
Levy          &  11          &  $[-10, 10]^{11}$       &  (1, $\dots$, 1)     & 0      \\
Rastrigin          &  11          &  $[-5.12, 5.12]^{11}$       &  (0, $\dots$, 0)     & 0      \\
Rosenbrock          &  11          &  $[-5, 10]^{11}$       &  (1, $\dots$, 1)      & 0      \\

Ackley          &  12          &  $[-10, 10]^{12}$       &   (0, $\dots$, 0)     & 0      \\
Levy          &  12          &  $[-10, 10]^{12}$       &  (1, $\dots$, 1)     & 0      \\
Rastrigin          &  12          &  $[-5.12, 5.12]^{12}$       &  (0, $\dots$, 0)     & 0      \\
Rosenbrock          &  12          &  $[-5, 10]^{12}$       &  (1, $\dots$, 1)      & 0      \\

Ackley          &  13          &  $[-10, 10]^{13}$       &   (0, $\dots$, 0)     & 0      \\
Levy          &  13          &  $[-10, 10]^{13}$       &  (1, $\dots$, 1)     & 0      \\
Rastrigin          &  13          &  $[-5.12, 5.12]^{13}$       &  (0, $\dots$, 0)     & 0      \\
Rosenbrock          &  13          &  $[-5, 10]^{13}$       &  (1, $\dots$, 1)      & 0      \\

Ackley          &  14          &  $[-10, 10]^{14}$       &   (0, $\dots$, 0)     & 0      \\
Levy          &  14          &  $[-10, 10]^{14}$       &  (1, $\dots$, 1)     & 0      \\
Rastrigin          &  14          &  $[-5.12, 5.12]^{14}$       &  (0, $\dots$, 0)     & 0      \\
Rosenbrock          &  14          &  $[-5, 10]^{14}$       &  (1, $\dots$, 1)      & 0      \\

Ackley          &  15          &  $[-10, 10]^{15}$       &   (0, $\dots$, 0)     & 0      \\
Levy          &  15          &  $[-10, 10]^{15}$       &  (1, $\dots$, 1)     & 0      \\
Rastrigin          &  15          &  $[-5.12, 5.12]^{15}$       &  (0, $\dots$, 0)     & 0      \\
Rosenbrock          &  15          &  $[-5, 10]^{15}$       &  (1, $\dots$, 1)      & 0      \\
\bottomrule
\label{tab:tab_c4_bo_benchmark_fns}
\end{tabular}
\end{footnotesize}
\end{table}

\textbf{Experiment protocol}.
For each benchmark function, all 88 BO algorithm variants were evaluated and each experiment was repeated five times using different random seeds. 
The reported results correspond to the mean and standard error across the five runs. 
Within each run, all algorithms were initialised using the same set of input points to ensure a fair comparison. 
Each algorithm was allocated a total evaluation budget of $50d$. 
Batches of query points were generated by greedily optimising the acquisition function, with query points selected in descending order of acquisition value. 
The evaluation of an entire batch was counted as a single BO iteration.

\textbf{Evaluation metrics}.
The optimisation performance of a BO algorithm after an evaluation budget of $T$ was measured using the simple regret $r_T$ \citep{Bubeck2009,Dorard2012,Shahriari2016}, defined as 
\[ 
r_T = \min_{\vx' \in \mathcal{D}_T} y(\vx') - \min_{\vx \in \mathcal{X}} f(\vx) 
\] 
that is, the difference between the objective value of the best queried point and that of the true global minimiser.
Simple regret was reported on a logarithmic scale.

For pairwise comparisons between BO algorithms, simple regrets were analysed using Student's $t$-test. 
The resulting $p$-values quantified the strength of evidence for performance differences between algorithms. Throughout this study, differences with $p \leq 0.1$ were regarded as statistically significant. 

Computational overhead was defined as the time required to select the next set of query points. This included both GP model fitting and acquisition-function optimisation. 

All experiments reporting computational overhead were conducted on a single Apple M2 processor (3.49 GHz, 8 GB memory).

\textbf{Presentation of results}.
To provide a comprehensive overview of optimisation performance and computational overhead, the results are analysed using 
(i) performance progress plots, 
(ii) computational overhead progress plots, 
(iii) performance-overhead trade-off plots, 
(iv) violin plots, 
(v) summary statistics tables, and 
(vi) $p$-value tables.

Performance progress plots present the average accumulated simple regret across five runs at each optimisation step. 
Computational overhead progress plots show the average optimisation overhead at each BO iteration. 
Performance-overhead trade-off plots compare final simple regret against average computational overhead over the entire optimisation run.
All simple regret values are reported on a logarithmic scale.

Violin plots illustrate the distribution of final simple regrets over five runs, with horizontal lines indicating the mean, median, minimum, and maximum values. The vertical axis corresponds to the final simple regret on a logarithmic scale.

Below each violin plot, the first table reports the mean, geometric mean, and median final simple regret. Shaded cells identify the three best-performing algorithms according to each metric.

The second table reports $p$-values obtained from paired $t$-tests comparing final simple regrets between algorithms. The ordering of rows and columns is identical. Each entry indicates that the algorithm corresponding to the column outperforms the algorithm corresponding to the row, whereas an empty cell indicates the opposite. Values with $p < 0.1$ are highlighted to indicate statistically significant differences.

\subsection{Comparative studies of Gaussian process models}

\begin{figure}
\centering
\includegraphics[width=1.0\textwidth]{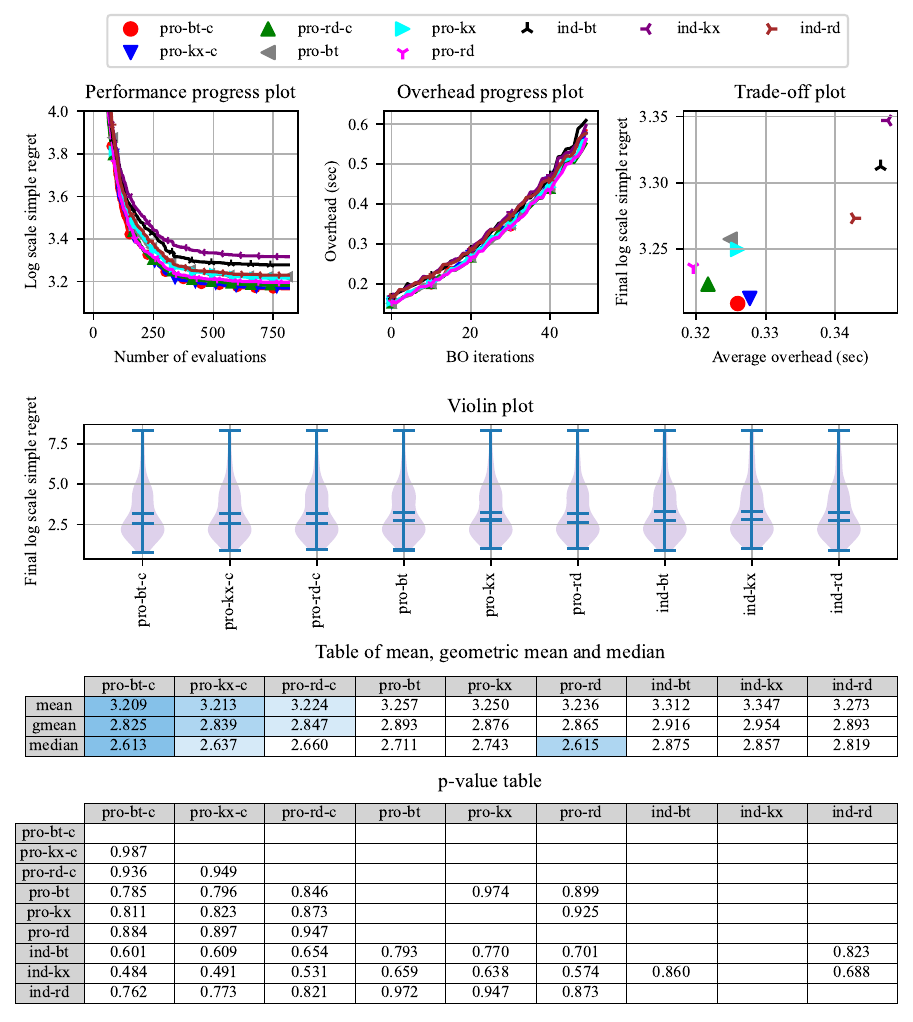}
\caption{ 
Performance and computational overhead on 47 synthetic benchmark functions across 12 input dimensions ($d = 2, 5, \dots, 15$, $B = d$), comparing three data assignment methods: random (rd), $k$-means (kx), and ball-tree (bt).
Results are averaged over all functions. Variants of BO-pro-c (pro-c), BO-pro (pro), and BO-ind (ind) are shown.
The performance progress plot shows the average accumulated simple regrets.
The overhead progress plot shows the average computational overheads.
The trade-off plot presents the comparison of the final simple regret with average computational overhead.
The violin plot gives the entire range of final simple regrets.
The first table presents the mean, geometric mean, and median of the final simple regrets. 
The second table presents $p$-values comparing final simple regrets between algorithms.
$p \leqslant 0.1$ indicates statistically significant differences. Each entry indicates that the column algorithm outperforms the row algorithm, while empty cells indicate the opposite.
BO-pro-c with ball-tree assignment (pro-bt-c) achieved the best overall performance.
}
\label{fig:fig_c4_regret_time_progress_violin_partition}
\end{figure}

Figure~\ref{fig:fig_c4_regret_time_progress_violin_partition} presents the performance and computational overhead of the BO-pro-c, BO-pro, and BO-ind algorithms across three data assignment methods: random (rd), $k$-means (kx), and ball-tree (bt). For this analysis, algorithms were grouped according to their data assignment method, and the reported performance and computational overhead were averaged across all algorithms sharing the same assignment method and across the 47 synthetic benchmark functions. 
All BO-pro-c and BO-pro variants employed entropy weighting and the GP-LCB acquisition function.

\begin{figure}
\centering
\includegraphics[width=1.0\textwidth]{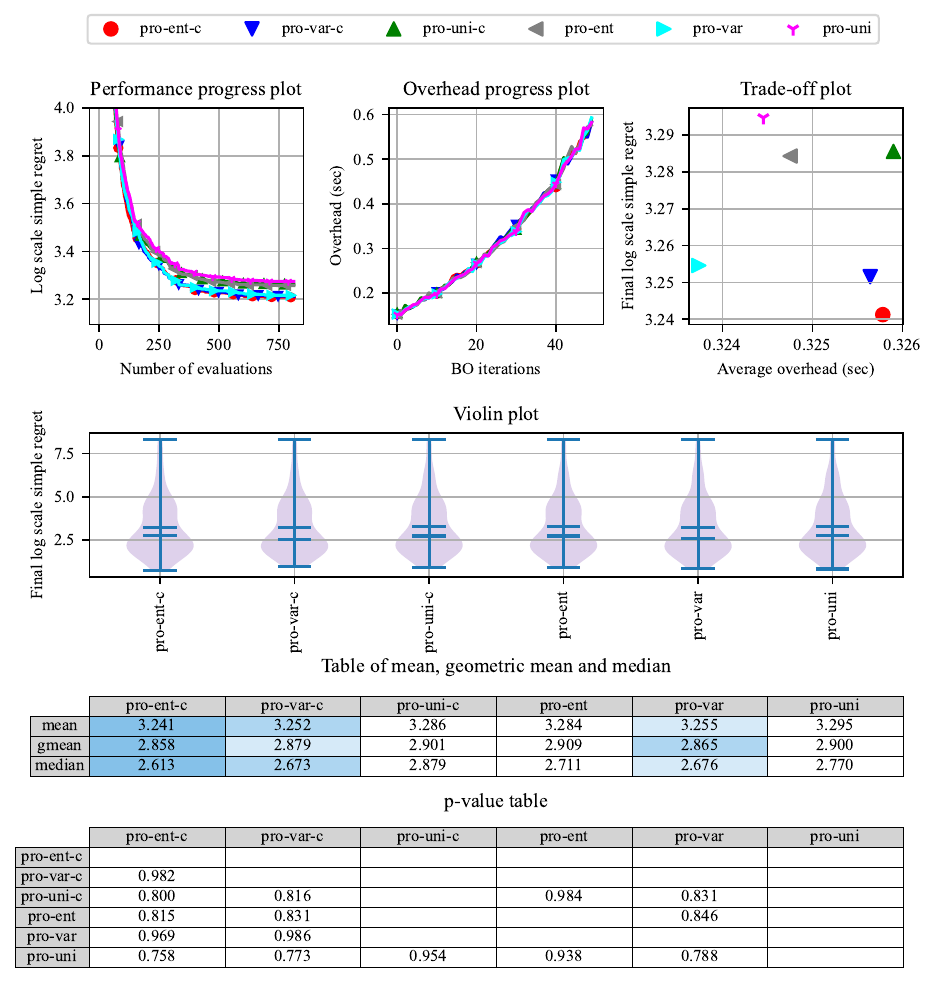}
\caption{ 
Performance and computational overhead on 47 synthetic benchmark functions across 12 input dimensions ($d = 2, 5, \dots, 15$, $B = d$), comparing three weighting schemes: entropy (ent), variance (var), and uniform (uni).
Results are averaged over all functions. Variants of BO-pro-c (pro-c) and BO-pro (pro) are shown.
BO-pro-c with entropy weighting (pro-ent-c) achieved the best overall performance.
}
\label{fig:fig_c4_regret_time_progress_violin_weight}
\end{figure}

Figure~\ref{fig:fig_c4_regret_time_progress_violin_weight} presents the performance and computational overhead of BO-pro-c and BO-pro across three weighting schemes: entropy (ent), variance (var), and uniform (uni). For this analysis, algorithms were grouped according to their weighting scheme, and the reported performance and computational overhead were averaged across all algorithms sharing the same weighting scheme and across the 47 synthetic benchmark functions. 
All BO-pro-c and BO-pro variants employed the ball-tree assignment method and the GP-LCB acquisition function.

In general, BO-pro-c variants achieved lower average simple regret than their BO-pro counterparts across all data assignment methods and weighting schemes, although none of these differences were statistically significant. These results suggest that information-based uncertainty calibration provides consistent performance improvements within a BO framework, regardless of the data assignment method or weighting scheme employed.

As shown in the performance progress plots in Figure~\ref{fig:fig_c4_regret_time_progress_violin_partition}, BO-pro-c generally converges more rapidly than BO-pro. This observation suggests that information-based uncertainty calibration helps maintain a more effective balance between exploration and exploitation throughout the optimisation process.

Among the BO-pro-c variants using random, $k$-means, and ball-tree assignment methods (denoted by pro-rd-c, pro-kx-c, and pro-bt-c, respectively), pro-bt-c achieved the best overall performance, attaining the lowest mean, geometric mean, and median simple regrets. However, none of the performance differences between the three assignment methods were statistically significant.

In terms of computational overhead, the $k$-means assignment method incurred slightly higher costs than the random and ball-tree methods. This behaviour can be attributed to the uneven allocation of data points across local GP models under $k$-means clustering. Consequently, some local models contained substantially more data points than others, resulting in higher computational costs. In contrast, both random and ball-tree assignment produced more balanced data partitions and therefore more consistent computational overhead.

Among the weighting schemes, BO-pro-c variants using entropy weighting achieved the best overall performance, yielding the lowest mean, geometric mean, and median simple regrets. The strong performance of the ball-tree assignment method and entropy-weighted aggregation is consistent with the findings reported in \cite{Ong2026}.

The computational overhead trends observed for BO-pro-c closely resemble those of BO-pro. This indicates that the uncertainty calibration procedure introduces only a marginal additional computational cost.

Figure~\ref{fig:fig_c4_regret_time_progress_violin_partition} also shows that, among the BO-ind variants (ind-rd, ind-kx, and ind-bt), ind-rd performed noticeably better than the other two approaches. 
This finding suggests that random partitioning is a more effective strategy for allocating training data to independent local GP models when no mechanism for information sharing exists between experts. 
A possible explanation is that random assignment reduces discontinuities in predictions at the boundaries between local regions. As observed for BO-pro-c and BO-pro, ind-kx also incurred slightly higher computational overhead than ind-rd and ind-bt due to the uneven distribution of data points among local GP models.

Comparing BO-pro-c, BO-pro, and BO-ind more broadly, all BO-pro-c and BO-pro variants achieved lower average simple regret than their BO-ind counterparts. Furthermore, BO-ind incurred higher computational overhead than both BO-pro-c and BO-pro. This is because each local GP model in BO-ind independently evaluates the acquisition function over the same candidate set at every optimisation step, causing computational cost to increase with the number of local models.

Overall, these comparative studies suggest that information-based uncertainty calibration can provide an effective mechanism for mitigating variance overestimation in product-of-experts GP models and improving BO performance. 
The strongest overall empirical performance was obtained when the GP-pro model was constructed using the ball-tree assignment method and aggregated using the entropy-weighted scheme.

\subsection{Examining the impact of the number of data points per local Gaussian process model}

In BO settings involving large-scale datasets, the number of data points assigned to each local GP model is an important factor influencing both optimisation performance and computational overhead. A larger local training set generally enables a GP model to produce more accurate predictions, albeit at the expense of increased computational cost.

\begin{figure}
\centering
\includegraphics[width=1.0\textwidth]{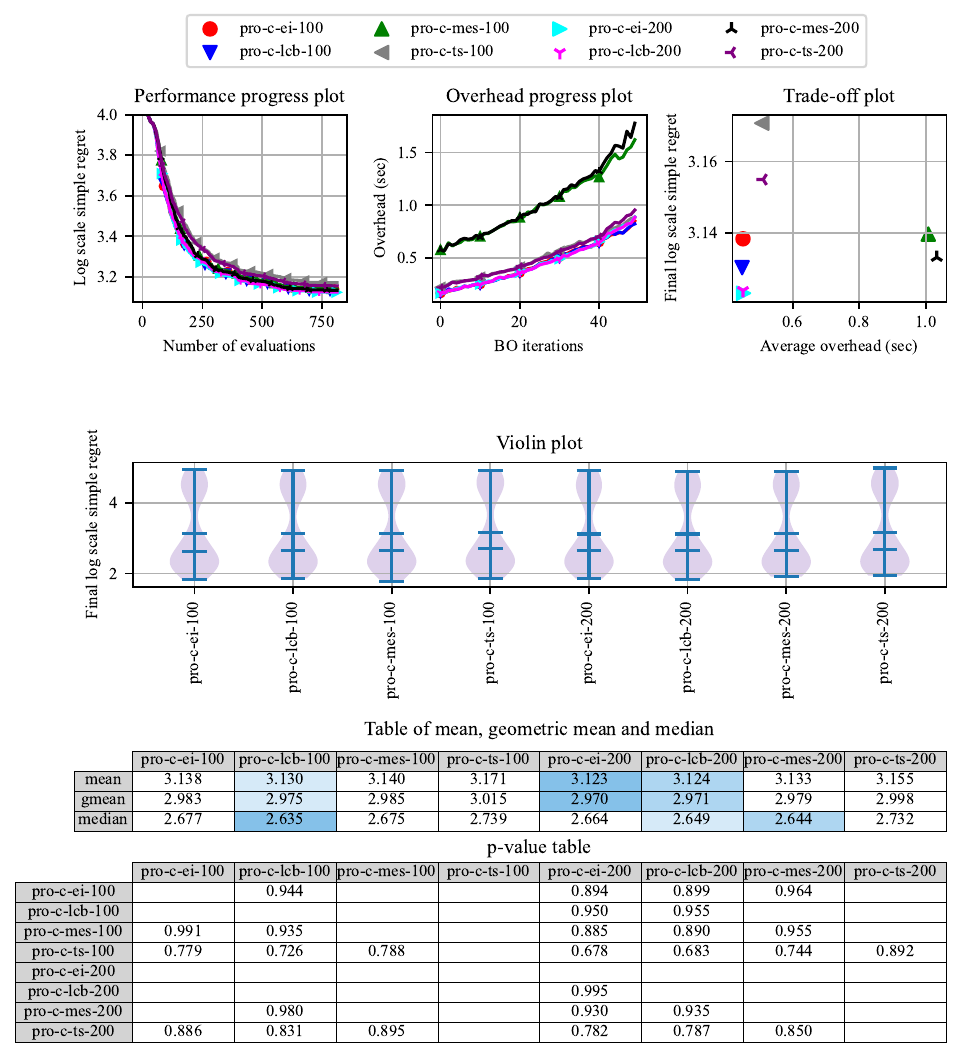}
\caption{ 
Performance and computational overhead on 18 synthetic benchmark functions across six input dimensions ($d = 10, \dots, 15$, $B = d$), comparing BO-pro-c with 100 and 200 data points per local GP model. Results are averaged across all benchmark functions and acquisition functions. BO-pro-c with 200 data points per local GP model (BO-pro-c-200) consistently achieved better optimisation performance than BO-pro-c-100, at the cost of a modest increase in computational overhead. }
\label{fig:fig_c4_regret_time_progress_violin_100}
\end{figure}

Figure~\ref{fig:fig_c4_regret_time_progress_violin_100} presents the performance and computational overhead of BO-pro-c using 100 and 200 data points per local GP model across four acquisition functions. 
Experiments were conducted on 18 synthetic benchmark functions, namely the 10d--15d Ackley, Levy, and Rastrigin functions, which require relatively large numbers of observations for effective optimisation.

For this analysis, algorithms were grouped according to the number of data points assigned to each local GP model and the acquisition function used. The reported results were averaged across the 18 benchmark functions. All BO-pro-c variants employed the ball-tree assignment method and the entropy-weighted aggregation scheme.

Across all acquisition functions, BO-pro-c variants using 200 data points per local GP model (denoted by BO-pro-c-200) generally achieved lower average simple regrets than those using 100 data points (denoted by BO-pro-c-100), although the improvement was accompanied by a modest increase in computational overhead. These findings suggest that larger local training sets enable the GP-pro-c model to approximate the underlying objective function more accurately, thereby supporting more effective exploration of the search space and improved optimisation performance.

Overall, these results indicate that, when sufficient computational resources are available, using a larger number of data points per local GP model is beneficial for BO-pro-c in large-scale optimisation settings.

\subsection{Comparative study of acquisition functions}
\label{sec:sec_c4_experiments_acq}

\begin{figure}
\centering
\includegraphics[width=1.0\textwidth]{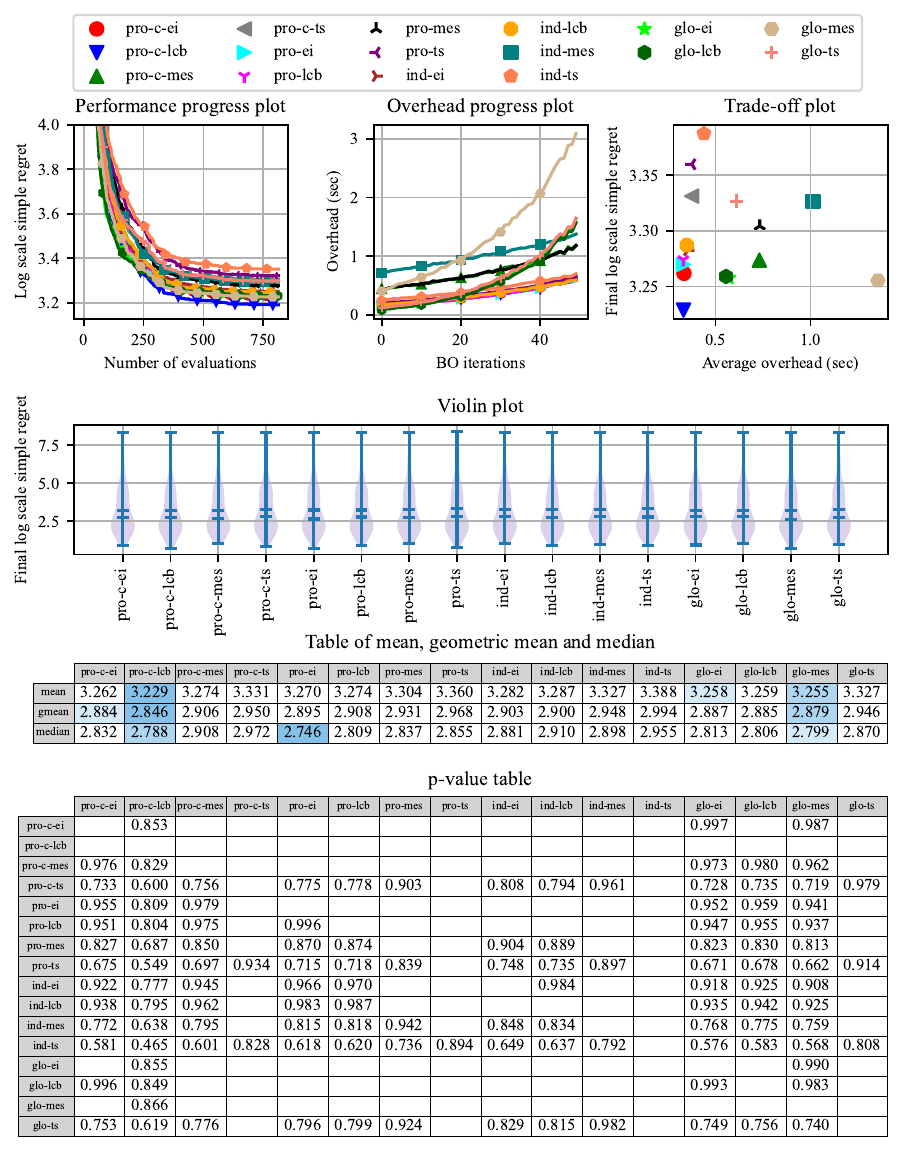}
\caption{ 
Performance and computational overhead on 47 synthetic benchmark functions across 12 input dimensions ($d = 2, 5, \dots, 15$, $B = d$), comparing four acquisition functions: EI, GP-LCB, MES, and TS. Results are averaged across all benchmark functions. BO-pro-c with GP-LCB (pro-c-lcb) achieved the best overall empirical performance. 
}
\label{fig:fig_c4_regret_time_progress_violin_acq}
\end{figure}

Figure~\ref{fig:fig_c4_regret_time_progress_violin_acq} presents the performance and computational overhead of BO-pro-c, BO-pro, BO-ind, and BO-glo across four acquisition functions: expected improvement (EI), Gaussian process lower confidence bound (GP-LCB), max-value entropy search (MES), and Thompson sampling (TS).

For this analysis, algorithms were grouped according to the acquisition function used, and the reported results were averaged across all algorithms sharing the same acquisition function and across the 47 benchmark functions. All BO-pro-c and BO-pro variants employed the ball-tree assignment method and the entropy-weighted aggregation scheme, whereas all BO-ind variants employed random data assignment.

Overall, BO-pro-c performed better than BO-pro and BO-ind across all acquisition functions. Specifically, the BO-pro-c variants (pro-c-ei, pro-c-lcb, pro-c-mes, and pro-c-ts) consistently achieved lower average simple regrets than the corresponding BO-pro and BO-ind variants.

Among all algorithms, BO-pro-c using the GP-LCB acquisition function exhibited the strongest overall empirical performance. The GP-LCB acquisition function guides the selection of query points using the predictive mean and confidence intervals provided by the GP surrogate model. Within BO-pro-c, the information-based calibration procedure reduces variance overestimation in local GP models, resulting in more accurate confidence intervals in the aggregated posterior distribution. Consequently, the acquisition function is less prone to excessive exploration of highly uncertain regions and is better able to balance exploration and exploitation throughout the optimisation process. This likely explains the superior convergence behaviour observed for BO-pro-c-lcb.

Although the empirical results suggest that GP-LCB is particularly effective when combined with BO-pro-c, no theoretical regret bounds are currently available for the proposed algorithm. The derivation of such guarantees remains an important direction for future work.

In terms of computational overhead, BO-pro-c, BO-pro, BO-ind, and BO-glo exhibited broadly similar computational trends when used with EI, GP-LCB, and TS. In contrast, the use of MES increased computational overhead by approximately one order of magnitude across all surrogate models. This additional cost is attributable to the approximation procedures required for evaluating the MES acquisition function.

Overall, these results indicate that BO-pro-c is most effective when combined with GP-LCB, providing the best optimisation performance while maintaining relatively low computational overhead.

\subsection{Examining the impact of input dimensionality}

\begin{figure}
\centering
\includegraphics[width=1.0\textwidth]{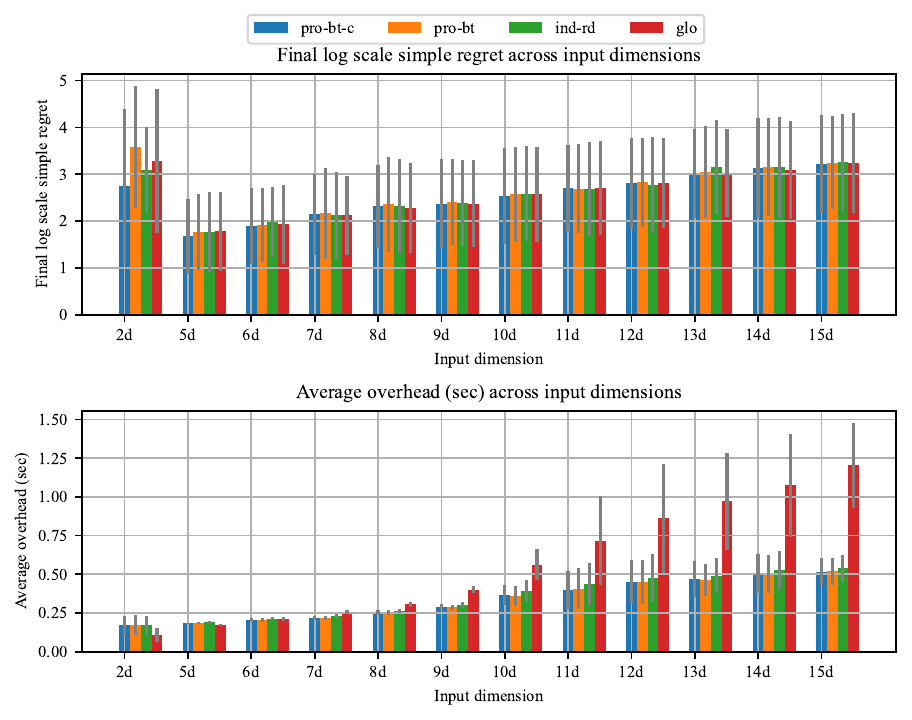}
\caption{ 
Performance and computational overhead on 47 synthetic benchmark functions across input dimensions $2 \leqslant d \leqslant 15$. The top plot shows the average simple regret, and the bottom plot shows the computational overhead. BO-pro-c, BO-pro, and BO-ind exhibited modest increases in overhead as dimensionality increases, whereas BO-glo exhibited a substantially steeper increase. 
}
\label{fig:fig_c4_regret_time_progress_violin_dim}
\end{figure}

Figure~\ref{fig:fig_c4_regret_time_progress_violin_dim} examines the effect of input dimensionality ($2 \leqslant d \leqslant 15$) on the performance and computational overhead of BO-pro-c, BO-pro, BO-ind, and BO-glo.

For this analysis, algorithms were grouped according to their GP model type, and the reported performance and computational overhead were averaged across the corresponding groups of benchmark functions. 
All BO-pro-c and BO-pro variants employed the ball-tree assignment method and the entropy-weighted aggregation scheme. 
All BO-ind variants employed random data assignment. 
All algorithms used the GP-LCB acquisition function.

Several findings can be drawn from Figure~\ref{fig:fig_c4_regret_time_progress_violin_dim}. 
Across all 12 input dimensionalities, BO-pro-c (pro-c) achieved lower simple regret than both BO-pro (pro) and BO-ind (ind). Relative to BO-glo, BO-pro-c achieved lower regret on 9 of the 12 dimensionalities considered, while performing slightly worse at $d = 7$, $9$, and $14$.

In terms of computational overhead, BO-pro-c, BO-pro, and BO-ind exhibited only modest increases as dimensionality increased. 
In contrast, the computational overhead of BO-glo increased substantially with growing dimensionality. 
For lower-dimensional problems ($2 \leqslant d \leqslant 6$), BO-glo incurred lower computational overhead than the other methods. However, for higher-dimensional problems ($7 \leqslant d \leqslant 15$), BO-glo became substantially more expensive than BO-pro-c, BO-pro, and BO-ind.

\section{Discussion and conclusion}
\label{sec:sec_s4_conclusion}

Table~\ref{tab:tab_c4_regret_all} reports the average log-scale simple regret achieved by BO-pro-bt-ent-c, BO-pro-bt-ent, BO-ind-rd, and BO-glo using the GP-LCB acquisition function across 47 synthetic benchmark functions. 
The corresponding average computational overhead per optimisation iteration is reported in Table~\ref{tab:tab_c4_overhead_all}.

Overall, the experimental results favour BO-pro-c, which incorporates an information-based uncertainty calibration mechanism within a product-of-experts GP model. The best overall empirical performance was achieved by BO-pro-c when combined with the ball-tree assignment method, entropy-based weighting, and the GP-LCB acquisition function.

Within BO-pro-c, the ball-tree assignment method groups spatially similar data points while maintaining balanced dataset sizes across local GP models, thereby improving local modelling accuracy without introducing substantial computational imbalance. 
The entropy-weighted aggregation scheme assigns larger weights to more reliable local experts and is closely related to minimising the KL divergence between individual expert distributions and the aggregated predictive distribution \citep{Cao2015}. 
Furthermore, the information-based calibration procedure reduces variance overestimation in local GP models, producing more accurate posterior uncertainty estimates. 
When combined with GP-LCB, these improved uncertainty estimates lead to more reliable confidence intervals and a better balance between exploration and exploitation during optimisation.

Compared with BO-pro, BO-pro-c achieved a 1.4\% reduction in average simple regret. 
These findings suggest that uncertainty calibration improves the quality of posterior uncertainty estimates and contributes to more effective optimisation behaviour.

Compared with BO-ind, BO-pro-c achieved a 1.8\% reduction in average simple regret and a 4.7\% reduction in computational overhead. This result is consistent with known limitations of independent local GP models, including discontinuities at sub-region boundaries \citep{Park2016}, the inability to capture long-range spatial correlations \citep{Liu2020}, and the lack of information sharing between local experts \citep{Liu2020}. Collectively, these limitations reduce the effectiveness of the optimisation process.

Compared with BO-glo, BO-pro-c achieved a modest 0.9\% reduction in average simple regret while reducing computational overhead by 39.4\%. These results suggest that BO-pro-c can retain optimisation performance comparable to that of a single global GP model while substantially reducing computational cost.

The relative advantages of BO-pro-c and BO-glo depend on problem dimensionality. 
For lower-dimensional problems ($d \leqslant 6$), BO-glo may be preferable because of its lower computational overhead while maintaining performance comparable to BO-pro-c. 
However, for higher-dimensional problems ($d > 6$), BO-pro-c generally provides a more favourable trade-off between optimisation performance and computational cost than either BO-glo or BO-ind.

Despite these improvements, BO-pro-c remains affected by the curse of dimensionality, particularly for problems with $d \geqslant 10$. Performance degradation in this regime is also observed for BO-glo and BO-ind. 
Future research could investigate the integration of BO-pro-c with trust-region methods to further improve scalability and optimisation performance in high-dimensional settings.

In summary, this study introduced BO-pro-c, a Bayesian optimisation algorithm that employs a product-of-experts Gaussian process model with uncertainty calibration as its surrogate model. Comprehensive experimental evaluations across 47 synthetic benchmark functions suggest that BO-pro-c can achieve competitive optimisation performance while maintaining substantially lower computational overhead than conventional BO approaches based on a single global GP model. These findings indicate that uncertainty-calibrated product-of-experts GP models provide an effective and scalable surrogate-modelling framework for Bayesian optimisation.

\begin{table}
\centering
\caption{Average log-scale simple regret of BO-pro-bt-ent-c, BO-pro-bt-ent, BO-ind-rd, and BO-glo using the GP-LCB acquisition function across 47 synthetic benchmark functions with input dimensionalities ranging from 2 to 15.}
\begin{footnotesize}
\begin{tabular}[t]{lccccl}
\toprule
Function & BO-pro-bt-ent-c-lcb &  BO-pro-bt-ent-lcb    &  BO-ind-rd-lcb   &  BO-glo-lcb  &  \\
\midrule
$2d$ Eggholder             &   $4.910 \pm 0.604$     &   $5.550 \pm 0.721$    &   $4.674 \pm 0.411$   &  $5.633 \pm 0.755$     \\
$2d$ Goldstein-Price     &   $2.436 \pm 1.430$     &   $2.613 \pm 1.159$    &   $2.728 \pm 0.718$   &  $1.842 \pm 0.483$     \\
$2d$ Shubert                 &   $3.613 \pm 1.644$     &   $4.475 \pm 0.603$    &   $3.355 \pm 0.986$   &  $4.830 \pm 0.165$     \\

$5d$ Ackley                 &   $1.597 \pm 0.057$     &   $1.678 \pm 0.040$    &   $1.702 \pm 0.082$   &  $1.640 \pm 0.155$    \\
$5d$ Levy                   &   $0.941 \pm 0.148$     &   $1.044 \pm 0.098$    &   $0.957 \pm 0.039$    &  $1.034 \pm 0.019$     \\
$5d$ Rastrigin             &   $2.861 \pm 0.126$     &   $2.991 \pm 0.191$    &   $3.005 \pm 0.090$   &  $3.030 \pm 0.084$    \\
$5d$ Rosenbrock        &   $4.504 \pm 0.384$   & $4.506 \pm 0.377$   & $5.788 \pm 0.664$   & $4.439 \pm 0.317$     \\

$6d$ Ackley                  &   $1.852 \pm 0.086$     &   $1.770 \pm 0.162$    &   $1.871 \pm 0.030$   &  $1.863 \pm 0.032$     \\
$6d$ Levy                     &   $1.134 \pm 0.107$     &   $1.255 \pm 0.059$    &   $1.316 \pm 0.073$   &  $1.144 \pm 0.032$     \\
$6d$ Rastrigin               &   $3.075 \pm 0.225$     &   $3.098 \pm 0.236$    &   $3.065 \pm 0.234$   &  $3.150 \pm 0.222$     \\
$6d$ Rosenbrock          &   $6.618 \pm 0.601$     &   $5.920 \pm 0.255$    &   $6.249 \pm 0.358$   &  $5.920 \pm 0.254$     \\

$7d$ Ackley                  &   $1.980 \pm 0.022$     &   $1.919 \pm 0.029$    &   $1.924 \pm 0.116$   &  $1.975 \pm 0.087$     \\
$7d$ Levy                     &   $1.459 \pm 0.023$     &   $1.477 \pm 0.029$    &   $1.447 \pm 0.151$   &  $1.396 \pm 0.052$     \\
$7d$ Rastrigin              &   $3.482 \pm 0.176$     &   $3.735 \pm 0.061$    &   $3.591 \pm 0.067$   &  $3.389 \pm 0.109$     \\
$7d$ Rosenbrock          &   $6.437 \pm 0.256$     &   $6.437 \pm 0.257$    &   $6.437 \pm 0.257$   &  $6.437 \pm 0.256$     \\

$8d$ Ackley                  &   $2.032 \pm 0.065$     &   $2.098 \pm 0.028$    &   $2.026 \pm 0.062$   &  $2.074 \pm 0.053$     \\
$8d$ Levy                     &   $1.745 \pm 0.110$     &   $1.651 \pm 0.112$    &   $1.682 \pm 0.103$   &  $1.557 \pm 0.219$     \\
$8d$ Rastrigin               &  $3.758 \pm 0.054$     &   $3.999 \pm 0.038$    &   $3.956 \pm 0.023$   &  $3.792 \pm 0.083$     \\
$8d$ Rosenbrock         &   $7.305 \pm 0.703$     &   $7.306 \pm 0.702$    &   $7.327 \pm 0.716$   &  $7.251 \pm 0.663$     \\

$9d$ Ackley                  &   $2.123 \pm 0.019$     &   $2.101 \pm 0.052$    &   $2.117 \pm 0.087$   &  $2.088 \pm 0.020$     \\
$9d$ Levy                     &   $1.719 \pm 0.019$     &   $1.867 \pm 0.021$    &   $1.755 \pm 0.035$   &  $1.795 \pm 0.054$     \\
$9d$ Rastrigin               &   $3.899 \pm 0.131$     &   $3.907 \pm 0.091$    &   $3.836 \pm 0.155$   &  $3.900 \pm 0.120$     \\
$9d$ Rosenbrock          &   $8.113 \pm 0.201$     &   $8.113 \pm 0.201$    &   $8.112 \pm 0.201$   &  $8.112 \pm 0.202$     \\

$10d$ Ackley                  &   $2.140 \pm 0.098$     &   $2.171 \pm 0.109$    &   $2.191 \pm 0.061$   &  $2.223 \pm 0.059$     \\
$10d$ Levy                     &   $2.058 \pm 0.060$     &   $2.037 \pm 0.088$    &   $2.050 \pm 0.155$   &  $2.000 \pm 0.117$     \\
$10d$ Rastrigin               &   $4.259 \pm 0.050$     &   $4.233 \pm 0.095$    &   $4.258 \pm 0.028$   &  $4.242 \pm 0.038$     \\
$10d$ Rosenbrock          &   $8.378 \pm 0.078$   & $8.378 \pm 0.078$   & $8.378 \pm 0.078$   & $8.378 \pm 0.078$     \\

$11d$ Ackley                  &   $2.207 \pm 0.050$     &   $2.197 \pm 0.066$    &   $2.184 \pm 0.049$   &  $2.189 \pm 0.094$     \\
$11d$ Levy                     &   $2.360 \pm 0.087$     &   $2.354 \pm 0.077$    &   $2.321 \pm 0.080$   &  $2.349 \pm 0.072$     \\
$11d$ Rastrigin               &   $4.207 \pm 0.098$     &   $4.269 \pm 0.171$    &   $4.373 \pm 0.112$   &  $4.376 \pm 0.110$     \\
$11d$ Rosenbrock          &   $8.957 \pm 0.061$   & $8.957 \pm 0.061$   & $8.957 \pm 0.061$   & $8.957 \pm 0.061$     \\

$12d$ Ackley                  &   $2.301 \pm 0.033$     &   $2.330 \pm 0.026$    &   $2.278 \pm 0.059$   &  $2.300 \pm 0.028$     \\
$12d$ Levy                     &   $2.438 \pm 0.107$     &   $2.450 \pm 0.121$    &   $2.382 \pm 0.133$   &  $2.460 \pm 0.129$     \\
$12d$ Rastrigin               &   $4.421 \pm 0.037$     &   $4.405 \pm 0.010$    &   $4.492 \pm 0.079$   &  $4.419 \pm 0.037$     \\
$12d$ Rosenbrock         &   $9.029 \pm 0.005$   & $9.029 \pm 0.005$   & $9.029 \pm 0.005$   & $9.029 \pm 0.005$     \\

$13d$ Ackley                  &   $2.306 \pm 0.018$     &   $2.308 \pm 0.019$    &   $2.278 \pm 0.073$   &  $2.325 \pm 0.040$     \\
$13d$ Levy                     &   $2.788 \pm 0.019$     &   $2.809 \pm 0.030$    &   $2.952 \pm 0.151$   &  $2.806 \pm 0.031$     \\
$13d$ Rastrigin               &   $4.496 \pm 0.196$     &   $4.597 \pm 0.065$    &   $4.636 \pm 0.087$   &  $4.500 \pm 0.189$     \\
$13d$ Rosenbrock         &   $9.965 \pm 0.067$   & $9.965 \pm 0.067$   & $9.965 \pm 0.067$   & $9.965 \pm 0.067$     \\

$14d$ Ackley                  &  $2.270 \pm 0.077$   & $2.275 \pm 0.052$   & $2.285 \pm 0.093$   & $2.246 \pm 0.077$    \\
$14d$ Levy                     &  $2.918 \pm 0.022$   & $2.955 \pm 0.019$   & $2.910 \pm 0.058$   & $2.879 \pm 0.063$     \\
$14d$ Rastrigin               &   $4.769 \pm 0.026$   & $4.755 \pm 0.054$   & $4.802 \pm 0.014$   & $4.724 \pm 0.030$     \\
$14d$ Rosenbrock         &   $9.809 \pm 0.352$   & $9.809 \pm 0.352$   & $9.809 \pm 0.352$   & $9.809 \pm 0.352$    \\

$15d$ Ackley                 &   $2.375 \pm 0.007$   & $2.392 \pm 0.010$   & $2.388 \pm 0.045$   & $2.377 \pm 0.007$     \\
$15d$ Levy                   &   $2.973 \pm 0.131$   & $3.051 \pm 0.083$   & $3.053 \pm 0.068$   & $3.005 \pm 0.153$     \\
$15d$ Rastrigin             &   $4.836 \pm 0.063$   & $4.748 \pm 0.069$   & $4.847 \pm 0.069$   & $4.861 \pm 0.081$    \\
$15d$ Rosenbrock       &   $10.585 \pm 0.194$   & $10.585 \pm 0.194$   & $10.585 \pm 0.194$   & $10.585 \pm 0.194$     \\
\bottomrule
Average $\downarrow$             &   \textbf{3.229}      &    3.274      &     3.287       &  3.259   \\
\bottomrule
\label{tab:tab_c4_regret_all}
\end{tabular}
\end{footnotesize}
\end{table}

\begin{table}
\centering
\caption{Average computational overhead per optimisation iteration (seconds) of BO-pro-bt-ent-c, BO-pro-bt-ent, BO-ind-rd, and BO-glo using the GP-LCB acquisition function across 47 synthetic benchmark functions with input dimensionalities ranging from 2 to 15.}
\begin{footnotesize}
\begin{tabular}[t]{lccccl}
\toprule
Function & BO-pro-bt-ent-c-lcb &  BO-pro-bt-ent-lcb    &  BO-ind-rd-lcb   &  BO-glo-lcb  &  \\
\midrule
$2d$ Eggholder                         &   $0.185 \pm 0.074$   & $0.188 \pm 0.075$   & $0.189 \pm 0.078$   & $0.113 \pm 0.044$     \\
$2d$ Goldstein-Price                  &   $0.146 \pm 0.001$   & $0.146 \pm 0.002$   & $0.147 \pm 0.002$   & $0.089 \pm 0.001$          \\
$2d$ Shubert                              &   $0.193 \pm 0.082$   & $0.193 \pm 0.085$   & $0.194 \pm 0.079$   & $0.123 \pm 0.060$          \\

$5d$ Ackley                &   $0.186 \pm 0.005$   & $0.186 \pm 0.003$   & $0.194 \pm 0.005$   & $0.173 \pm 0.008$    \\
$5d$ Levy                   &   $0.181 \pm 0.003$   & $0.179 \pm 0.004$   & $0.185 \pm 0.005$   & $0.170 \pm 0.009$    \\
$5d$ Rastrigin             &   $0.184 \pm 0.005$   & $0.184 \pm 0.006$   & $0.189 \pm 0.006$   & $0.174 \pm 0.006$     \\
$5d$ Rosenbrock        &   $0.190 \pm 0.016$   & $0.187 \pm 0.014$   & $0.194 \pm 0.016$   & $0.174 \pm 0.011$    \\

$6d$ Ackley                  &   $0.208 \pm 0.011$   & $0.214 \pm 0.008$   & $0.218 \pm 0.009$   & $0.223 \pm 0.013$    \\
$6d$ Levy                     &   $0.200 \pm 0.013$   & $0.199 \pm 0.011$   & $0.209 \pm 0.013$   & $0.203 \pm 0.009$     \\
$6d$ Rastrigin               &   $0.205 \pm 0.007$   & $0.206 \pm 0.010$   & $0.215 \pm 0.009$   & $0.205 \pm 0.006$     \\
$6d$ Rosenbrock          &   $0.196 \pm 0.002$   & $0.196 \pm 0.002$   & $0.205 \pm 0.004$   & $0.205 \pm 0.005$     \\

$7d$ Ackley                  &   $0.218 \pm 0.007$   & $0.217 \pm 0.009$   & $0.229 \pm 0.013$   & $0.256 \pm 0.015$     \\
$7d$ Levy                     &   $0.225 \pm 0.015$   & $0.222 \pm 0.013$   & $0.230 \pm 0.011$   & $0.260 \pm 0.011$    \\
$7d$ Rastrigin              &   $0.213 \pm 0.009$   & $0.220 \pm 0.013$   & $0.227 \pm 0.012$   & $0.251 \pm 0.008$    \\
$7d$ Rosenbrock          &   $0.222 \pm 0.004$   & $0.218 \pm 0.007$   & $0.229 \pm 0.008$   & $0.261 \pm 0.005$     \\\

$8d$ Ackley                  &   $0.251 \pm 0.014$   & $0.251 \pm 0.010$   & $0.263 \pm 0.012$   & $0.303 \pm 0.011$    \\
$8d$ Levy                     &   $0.253 \pm 0.019$   & $0.255 \pm 0.022$   & $0.267 \pm 0.016$   & $0.311 \pm 0.019$    \\
$8d$ Rastrigin              &   $0.252 \pm 0.013$   & $0.249 \pm 0.007$   & $0.261 \pm 0.012$   & $0.304 \pm 0.013$    \\
$8d$ Rosenbrock         &   $0.249 \pm 0.015$   & $0.253 \pm 0.020$   & $0.264 \pm 0.020$   & $0.301 \pm 0.017$     \\

$9d$ Ackley                  &   $0.293 \pm 0.015$   & $0.293 \pm 0.019$   & $0.306 \pm 0.019$   & $0.403 \pm 0.021$    \\
$9d$ Levy                     &   $0.296 \pm 0.012$   & $0.288 \pm 0.014$   & $0.301 \pm 0.011$   & $0.409 \pm 0.011$     \\
$9d$ Rastrigin               &   $0.284 \pm 0.018$   & $0.282 \pm 0.015$   & $0.298 \pm 0.019$   & $0.388 \pm 0.025$    \\
$9d$ Rosenbrock        &   $0.263 \pm 0.017$   & $0.259 \pm 0.017$   & $0.284 \pm 0.017$   & $0.362 \pm 0.021$     \\

$10d$ Ackley                  &   $0.353 \pm 0.052$   & $0.351 \pm 0.050$   & $0.377 \pm 0.060$   & $0.542 \pm 0.082$     \\
$10d$ Levy                     &   $0.357 \pm 0.050$   & $0.348 \pm 0.044$   & $0.380 \pm 0.047$   & $0.566 \pm 0.090$     \\
$10d$ Rastrigin               &   $0.387 \pm 0.088$   & $0.381 \pm 0.083$   & $0.418 \pm 0.089$   & $0.579 \pm 0.121$     \\
$10d$ Rosenbrock          &   $0.293 \pm 0.002$   & $0.283 \pm 0.001$   & $0.308 \pm 0.009$   & $0.427 \pm 0.003$     \\

$11d$ Ackley                  &   $0.345 \pm 0.042$   & $0.347 \pm 0.052$   & $0.377 \pm 0.056$   & $0.581 \pm 0.070$     \\
$11d$ Levy                     &   $0.448 \pm 0.154$   & $0.468 \pm 0.175$   & $0.497 \pm 0.168$   & $0.856 \pm 0.375$     \\
$11d$ Rastrigin               &   $0.407 \pm 0.122$   & $0.407 \pm 0.115$   & $0.434 \pm 0.123$   & $0.710 \pm 0.257$    \\
$11d$ Rosenbrock          &   $0.319 \pm 0.006$   & $0.316 \pm 0.011$   & $0.345 \pm 0.002$   & $0.536 \pm 0.009$     \\

$12d$ Ackley                  &   $0.503 \pm 0.165$   & $0.507 \pm 0.184$   & $0.536 \pm 0.189$   & $1.007 \pm 0.466$     \\
$12d$ Levy                     &   $0.396 \pm 0.060$   & $0.400 \pm 0.066$   & $0.414 \pm 0.071$   & $0.753 \pm 0.196$     \\
$12d$ Rastrigin               &   $0.457 \pm 0.145$   & $0.441 \pm 0.127$   & $0.486 \pm 0.154$   & $0.825 \pm 0.286$     \\
$12d$ Rosenbrock         &   $0.335 \pm 0.006$   & $0.337 \pm 0.013$   & $0.355 \pm 0.009$   & $0.602 \pm 0.027$    \\

$13d$ Ackley                  &   $0.464 \pm 0.094$   & $0.464 \pm 0.085$   & $0.482 \pm 0.080$   & $0.994 \pm 0.300$     \\
$13d$ Levy                     &   $0.430 \pm 0.044$   & $0.428 \pm 0.043$   & $0.451 \pm 0.042$   & $0.856 \pm 0.114$     \\
$13d$ Rastrigin               &   $0.517 \pm 0.166$   & $0.502 \pm 0.145$   & $0.540 \pm 0.161$   & $1.063 \pm 0.410$     \\
$13d$ Rosenbrock         &    $0.380 \pm 0.003$   & $0.386 \pm 0.004$   & $0.409 \pm 0.005$   & $0.748 \pm 0.007$     \\

$14d$ Ackley                  &   $0.481 \pm 0.062$   & $0.463 \pm 0.063$   & $0.506 \pm 0.061$   & $0.998 \pm 0.149$     \\
$14d$ Levy                     &   $0.562 \pm 0.189$   & $0.562 \pm 0.188$   & $0.588 \pm 0.186$   & $1.239 \pm 0.499$     \\
$14d$ Rastrigin               &   $0.471 \pm 0.043$   & $0.476 \pm 0.034$   & $0.494 \pm 0.048$   & $0.987 \pm 0.134$     \\
$14d$ Rosenbrock         &   $0.402 \pm 0.003$   & $0.401 \pm 0.003$   & $0.430 \pm 0.001$   & $0.830 \pm 0.001$     \\

$15d$ Ackley                 &   $0.530 \pm 0.094$   & $0.513 \pm 0.079$   & $0.544 \pm 0.098$   & $1.158 \pm 0.164$    \\
$15d$ Levy                   &   $0.502 \pm 0.092$   & $0.526 \pm 0.114$   & $0.535 \pm 0.096$   & $1.265 \pm 0.386$     \\
$15d$ Rastrigin             &   $0.519 \pm 0.078$   & $0.516 \pm 0.065$   & $0.540 \pm 0.063$   & $1.193 \pm 0.213$     \\
$15d$ Rosenbrock       &   $0.446 \pm 0.011$   & $0.425 \pm 0.003$   & $0.470 \pm 0.001$   & $0.988 \pm 0.010$    \\
\bottomrule
Average $\downarrow$             &  \textbf{0.323}       &   \textbf{0.322}       &   0.339         &  0.533   \\
\bottomrule
\label{tab:tab_c4_overhead_all}
\end{tabular}
\end{footnotesize}
\end{table}

\bibliographystyle{unsrtnat}
\bibliography{references}  

@article{Ong2026,
  author    = {Ong, Yean Hoon and Barucca, Paolo and Pan, Wei and Wang, Jun},
  title     = {Information-Based Calibration of Uncertainty Quantification in Product-of-Experts Gaussian Process Models},
  journal   = {Journal of Artificial Intelligence Research},
  volume    = {86},
  year      = {2026},
  publisher = {AI Access Foundation},
  doi       = {10.1613/jair.1.20374},
  url       = {https://doi.org/10.1613/jair.1.20374},
  month     = aug
}

@article{Brochu2010,
  author        = {Brochu, Eric and Cora, Vlad M. and de Freitas, Nando},
  title         = {A Tutorial on Bayesian Optimization of Expensive Cost Functions, with Application to Active User Modeling and Hierarchical Reinforcement Learning},
  journal       = {arXiv preprint arXiv:1012.2599},
  year          = {2010},
  eprint        = {1012.2599},
  archivePrefix = {arXiv},
  primaryClass  = {cs.LG},
  doi           = {10.48550/arXiv.1012.2599}
}

@book{garnett_bayesoptbook_2023,
  author    = {Garnett, Roman},
  title     = {Bayesian Optimization},
  year      = {2023},
  publisher = {Cambridge University Press}
}

@article{Frazier2018,
  author        = {Frazier, Peter I.},
  title         = {A Tutorial on Bayesian Optimization},
  journal       = {arXiv preprint arXiv:1807.02811},
  year          = {2018},
  eprint        = {1807.02811},
  archivePrefix = {arXiv},
  primaryClass  = {stat.ML},
  doi           = {10.48550/arXiv.1807.02811}
}

@article{Shahriari2016p,
  author  = {Shahriari, Bobak and Swersky, Kevin and Wang, Ziyu and Adams, Ryan P. and de Freitas, Nando},
  title   = {Taking the Human Out of the Loop: A Review of Bayesian Optimization},
  journal = {Proceedings of the IEEE},
  year    = {2016},
  volume  = {104},
  number  = {1},
  pages   = {148--175},
  doi     = {10.1109/JPROC.2015.2494218}
}

@book{Rasmussen2006,
  author    = {Rasmussen, Carl Edward and Williams, Christopher K. I.},
  title     = {Gaussian Processes for Machine Learning},
  year      = {2006},
  publisher = {MIT Press}
}

@inproceedings{Eriksson2019,
  author    = {Eriksson, David and Pearce, Michael and Gardner, Jacob and Turner, Ryan D. and Poloczek, Matthias},
  title     = {Scalable Global Optimization via Local Bayesian Optimization},
  booktitle = {Advances in Neural Information Processing Systems 32},
  year      = {2019},
  volume    = {32},
  publisher = {Curran Associates, Inc.}
}

@inproceedings{Schilling2016,
  author    = {Schilling, Nicolas and Wistuba, Martin and Schmidt-Thieme, Lars},
  title     = {Scalable Hyperparameter Optimization with Products of Gaussian Process Experts},
  booktitle = {Machine Learning and Knowledge Discovery in Databases},
  year      = {2016},
  pages     = {33--48},
  publisher = {Springer International Publishing},
  address   = {Cham},
  doi       = {10.1007/978-3-319-46128-1_3}
}

@article{Tautvaisas2022,
  author  = {Tautvai{\v{s}}as, Saulius and {\v{Z}}ilinskas, Julius},
  title   = {Scalable Bayesian Optimization with Generalized Product of Experts},
  journal = {Journal of Global Optimization},
  year    = {2022},
  volume  = {84},
  pages   = {217--238},
  doi     = {10.1007/s10898-022-01236-x}
}

@article{Jones1998,
  author  = {Jones, Donald R. and Schonlau, Matthias and Welch, William J.},
  title   = {Efficient Global Optimization of Expensive Black-Box Functions},
  journal = {Journal of Global Optimization},
  year    = {1998},
  volume  = {13},
  number  = {4},
  pages   = {455--492},
  doi     = {10.1023/A:1008306431147}
}

@article{Bull2011,
  author  = {Bull, Adam D.},
  title   = {Convergence Rates of Efficient Global Optimization Algorithms},
  journal = {Journal of Machine Learning Research},
  year    = {2011},
  volume  = {12},
  number  = {88},
  pages   = {2879--2904}
}

@inproceedings{balandat2020botorch,
  author    = {Balandat, Maximilian and Karrer, Brian and Jiang, Daniel R. and Daulton, Samuel and Letham, Benjamin and Wilson, Andrew Gordon and Bakshy, Eytan},
  title     = {BoTorch: A Framework for Efficient Monte-Carlo Bayesian Optimization},
  booktitle = {Advances in Neural Information Processing Systems 33},
  year      = {2020},
  volume    = {33}
}

@inproceedings{Gardner2018,
  author    = {Gardner, Jacob R. and Pleiss, Geoff and Bindel, David and Weinberger, Kilian Q. and Wilson, Andrew Gordon},
  title     = {GPyTorch: Blackbox Matrix-Matrix Gaussian Process Inference with GPU Acceleration},
  booktitle = {Advances in Neural Information Processing Systems 31},
  year      = {2018},
  volume    = {31},
  pages     = {7576--7587}
}

@inproceedings{Bubeck2009,
  author    = {Bubeck, S{\'e}bastien and Munos, R{\'e}mi and Stoltz, Gilles},
  title     = {Pure Exploration in Multi-Armed Bandits Problems},
  booktitle = {Algorithmic Learning Theory},
  year      = {2009},
  pages     = {23--37},
  publisher = {Springer Berlin Heidelberg},
  address   = {Berlin, Heidelberg},
  doi       = {10.1007/978-3-642-04414-4_5}
}

@phdthesis{Dorard2012,
  author = {Dorard, Louis},
  title  = {Bandit Algorithms for Searching Large Spaces},
  school = {University College London},
  year   = {2012}
}

@phdthesis{Shahriari2016,
  author = {Shahriari, Bobak},
  title  = {Practical Bayesian Optimization with Application to Tuning Machine Learning Algorithms},
  school = {The University of British Columbia},
  year   = {2016}
}

@article{Cao2015,
  author        = {Cao, Yanshuai and Fleet, David J.},
  title         = {Transductive Log Opinion Pool of Gaussian Process Experts},
  journal       = {arXiv preprint arXiv:1511.07551},
  year          = {2015},
  eprint        = {1511.07551},
  archivePrefix = {arXiv},
  primaryClass  = {cs.LG},
  doi           = {10.48550/arXiv.1511.07551}
}

@article{Park2016,
  author  = {Park, Chiwoo and Huang, Jianhua Z.},
  title   = {Efficient Computation of Gaussian Process Regression for Large Spatial Data Sets by Patching Local Gaussian Processes},
  journal = {Journal of Machine Learning Research},
  year    = {2016},
  volume  = {17},
  number  = {174},
  pages   = {1--29}
}

@article{Liu2020,
  author  = {Liu, Haitao and Ong, Yew-Soon and Shen, Xiaobo and Cai, Jianfei},
  title   = {When Gaussian Process Meets Big Data: A Review of Scalable GPs},
  journal = {IEEE Transactions on Neural Networks and Learning Systems},
  year    = {2020},
  volume  = {31},
  number  = {11},
  pages   = {4405--4423},
  doi     = {10.1109/TNNLS.2019.2957109}
}

\end{document}